\documentclass{midl} % Include author names
\newcommand{\Enzo}[1]{\textcolor{black}{#1}}

\usepackage{dsfont}
 \usepackage{booktabs}
 \usepackage{multirow}
 \usepackage{graphicx}
\usepackage{mwe} % to get dummy images
\usepackage{float} % to get dummy images

\jmlrvolume{}
\jmlryear{2019}
\jmlrworkshop{Full Paper -- MIDL 2019}
\editors{Accepted at MIDL 2019}

\title[Joint Learning of Brain Lesion and Anatomy Segmentation from Het. Datasets]{Joint Learning of Brain Lesion and Anatomy Segmentation from Heterogeneous Datasets}

\midlauthor{\Name{Nicolas Roulet\nametag{$^{1}$}} \Email{nroulet@dc.uba.ar}\\
\Name{Diego Fernandez Slezak\nametag{$^{1}$}} \Email{dfslezak@dc.uba.ar }\\
\Name{Enzo Ferrante\nametag{$^{2}$}} \Email{eferrante@sinc.unl.edu.ar}\\
\addr $^{1}$ Universidad de Buenos Aires, CONICET, Buenos Aires, Argentina \\
\addr $^{2}$ Universidad Nacional del Litoral, CONICET, Santa Fe, Argentina 
}
\begin{document}

\maketitle

\begin{abstract}
Brain lesion and anatomy segmentation in magnetic resonance images are fundamental tasks in neuroimaging research and clinical practice. Given enough training data, convolutional neuronal networks (CNN) proved to outperform all existent techniques in both tasks independently. However, to date, little work has been done regarding simultaneous learning of brain lesion and anatomy segmentation from disjoint datasets.

In this work we focus on training a single CNN model to predict brain tissue and lesion segmentations using heterogeneous datasets labeled independently, according to only one of these tasks (a common scenario when using publicly available datasets). We show that label contradiction issues can arise in this case, and propose a novel \textit{adaptive cross entropy} (ACE) loss function that makes such training possible. We provide quantitative evaluation in two different scenarios, benchmarking the proposed method in comparison with a multi-network approach. Our experiments suggest ACE loss enables training of single models when standard cross entropy and Dice loss functions tend to fail. Moreover, we show that it is possible to achieve competitive results when comparing with multiple networks trained for independent tasks.

\end{abstract}

\begin{figure}[t!]
\includegraphics[width=1\linewidth]{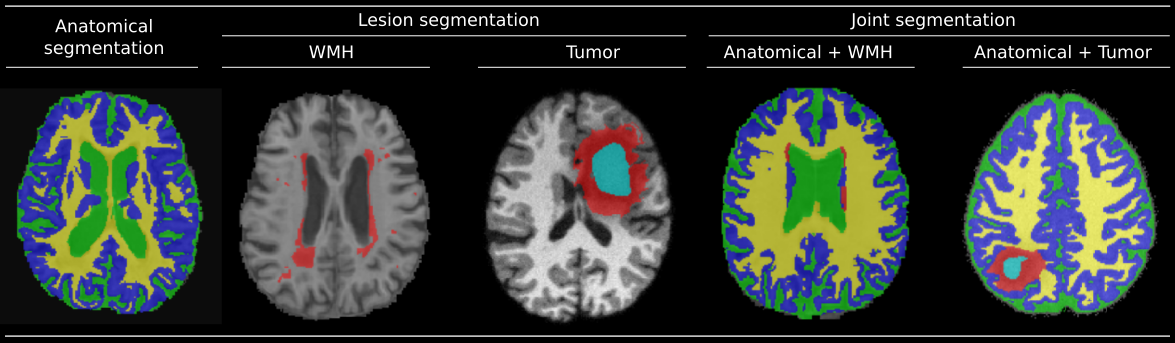}
\caption{Example of brain MRI with overlapped annotations corresponding to anatomy, lesion and joint segmentations. Note that whatever is considered as background in both, WMH and tumor segmentation datasets, should be classified as tissue according to the anatomy dataset. This fact misleads the training process of a single CNN when using standard categorical cross entropy or Dice losses to perform joint learning of lesion and anatomy segmentation.}
\label{fig:intro}
\end{figure}

\section{Introduction}
% Papers importantes: 
% - https://arxiv.org/pdf/1802.04962.pdf
% - https://arxiv.org/pdf/1802.06664.pdf

Segmentation of anatomical and pathological structures in volumetric images is a fundamental task for biomedical image analysis. It constitutes the first step in several medical procedures such as shape analysis for population studies, computed assisted diagnosis/surgery and automatic radiotherapy planning, among many others. Segmentation accuracy is therefore of paramount importance in these cases, since it will necessarily influence the overall quality of such procedures.

During the last years, convolutional neural networks (CNNs) proved to be highly accurate to perform medical image segmentation \cite{unet,deepmedic,kamnitsas2017ensembles,shakeri2016sub}. In this scenario, a training dataset consists of medical images with expert annotations associated to a particular task of interest. Following a supervised approach, CNNs are trained to perform such task by learning the network parameters that minimize a given loss function over the training data. In the context of brain image segmentation (of main interest in this work), publicly available datasets with manual annotations usually correspond to single tasks. These tasks might be associated to anatomy segmentation (e.g. brain tissues \cite{mendrik2015mrbrains,cocosco1997brainweb},  sub-cortical structures \cite{ibsr}) or pathological segmentation (e.g. brain tumours \cite{brats12}, white matter hiper-intensities \cite{mrbrains17}).

Even if most publicly available datasets provide image annotations for single tasks, in practice it is usually desirable to train single models which can learn to perform multiple segmentation tasks simultaneously. We focus on the particular case of brain magnetic resonance images (MRI), where segmenting both brain lesions and anatomical structures is especially relevant. For example, in the context of neurovascular and neurodegenerative diseases \cite{moeskops2018evaluation}, white matter hyper-intensity (WMH) segmentation in brain MRI is usually combined with brain tissue segmentation when studying cognitive dysfunction in elderly patients \cite{de2010progression}. Another example is related to brain tumour segmentation \cite{menze2015multimodal}. Combining brain tumor segmentation with brain tissue classification \cite{moon2002} would have an enormous potential value for improved medical research and biomarkers discovery. We will explore both application scenarios and provide experimental evidence about the effectiveness of the proposed method to perform joint learning of brain lesion and anatomy segmentation in these cases.

Learning to segment multiple structures from heterogeneous datasets is a challenging task, since labels coming from different datasets may contradict each other and mislead the training process. In the particular case of brain lesion and anatomy segmentation from MRI, Figure \ref{fig:intro} illustrates this issue. Given two datasets with disjoint labels (for example, brain tissues and WMH lesions), whatever is considered as background in the lesion dataset, should be classified as tissue according to the anatomy dataset. This raises a label contradiction problem that will be studied in this work.

We interpret brain lesion and anatomy segmentation as two different tasks which are learned from heterogeneous datasets, meaning that each dataset is annotated for a single task. In what follows, we briefly describe related works about learning to segment from disjoint annotations, discuss the issues that arise when training a single CNN model to perform both tasks with standard loss functions, and propose a simple, yet effective, adaptive loss function that makes it possible to train such model using heterogeneous datasets.

\subsection{Related Work}
\label{sec:soa}

Similar multi-task problems in the context of image segmentation were explored in recent works. Regarding segmentation for medical images,  \cite{moeskops2016deep} studied how a single deep CNN can be used to predict multiple anatomical structures for three different tasks including brain MRI, breast MRI and cardiac computed tomography angiography (CTA) segmentation. They showed that a standard combined training procedure with balanced mini-batch sampling results in segmentation performance equivalent to that of a deep CNN trained specifically for that task. This problem differs from our setting since every dataset is associated to a different organ. Therefore, labels from different datasets can not co-exists in a single image avoiding the label contradiction problem illustrated in Figure \ref{fig:intro}. 

Closest to our work are those by \cite{fourure2017,rajchl2018neuronet}, where a single segmentation model is learned from multiple training datasets defined on images representing similar domains. In \cite{fourure2017}, the authors train a model to perform semantic full scene labeling in outdoor images coming from different datasets with heterogeneous labels. They propose a \textit{selective cross entropy} loss that, instead of considering a single final \textit{softmax} activation function defined over the entire set of possible labels, is computed using a dataset-wise \textit{softmax} activation function. This dataset-wise \textit{softmax} only takes into account those labels available in the dataset corresponding to the current training sample. A similar strategy is followed by \cite{rajchl2018neuronet} in the context of brain image segmentation. The authors propose the NeuroNet, a multi-output CNN that mimics several popular and state-of-the-art brain segmentation tools producing segmentations for brain tissues, cortical and sub-cortical structures. Differently from \cite{fourure2017}, NeuroNet combines a multi-decoder architecture (one decoder for every dataset/task) with an analogous multi-task loss based on cross entropy, defined as the average of independent loss functions computed for every single task. Note that our problem differs from those tackled in both papers: our aim is to produce a segmentation model that assigns a single label to every voxel (considering the union of anatomical and pathological labels). On the contrary, they aim at predicting one and exactly one label from each labelset for every voxel, i.e. multiple labels will be assigned to every voxel. %Inspired by these previous works, we propose an \textit{adaptive cross entropy} loss (ACE) that enables joint learning of brain lesion and anatomy segmentation from disjoint datasets.
\begin{figure}[t!]
\includegraphics[width=1\linewidth]{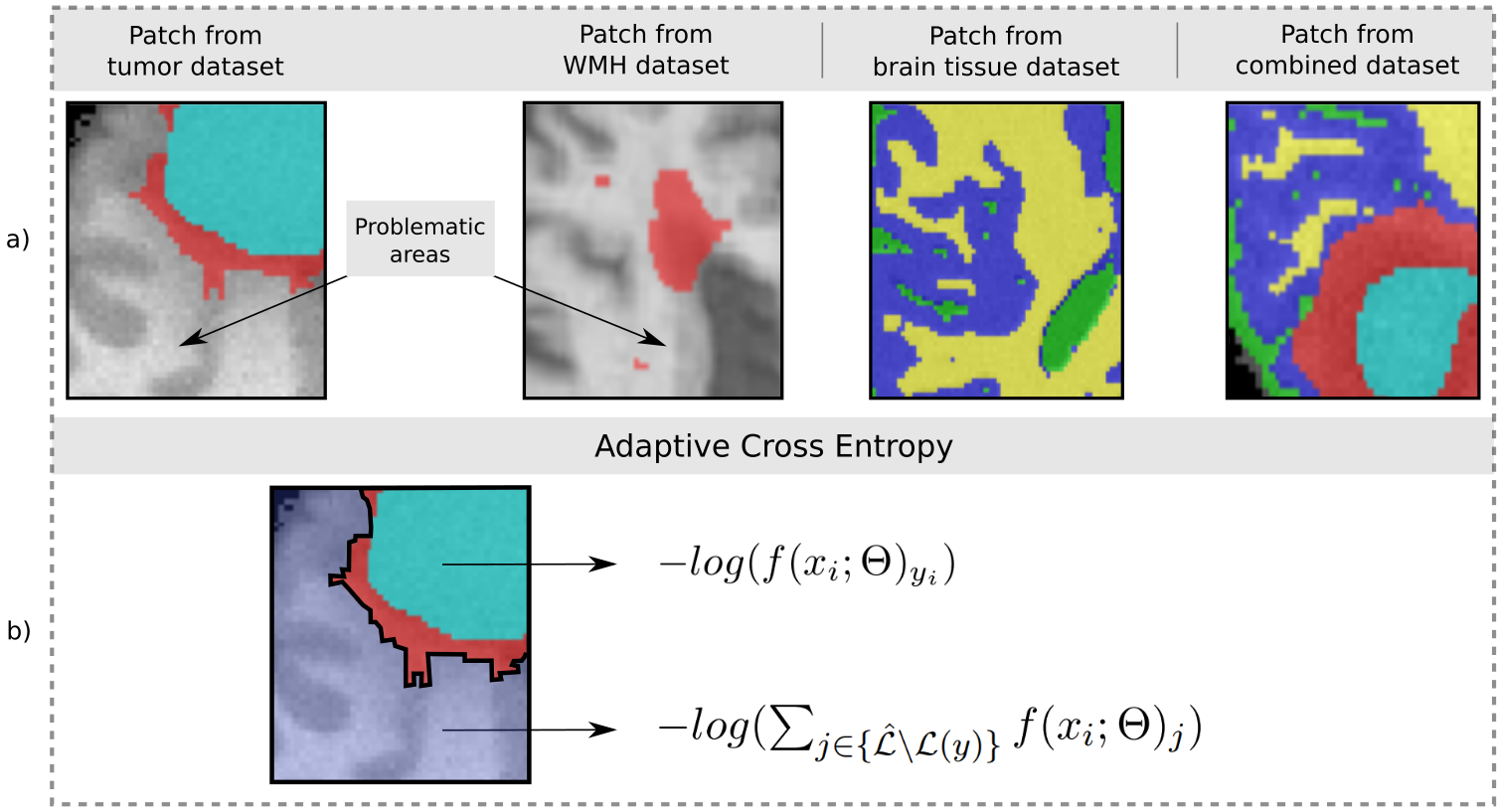}
\caption{(a) Example of image patches with overlapped segmentation masks sampled from: the lesion datasets (tumor and WMH), the anatomical (brain tissue) dataset and the desired combined segmentation for which we do not have training data. Problematic areas are those for which the original lesion datasets indicate background label, while they should be annotated as actual tissue labels.\newline (b) The proposed \textit{adaptive cross entropy} behaves differently depending on the structures of interest under consideration. We reinterpret the meaning assigned to the lesion background label (in blue) as \textit{'any label that is not lesion'} and modify the loss function accordingly.}
\label{fig:ace}
\end{figure}

\section{Learning Brain Lesion and Anatomy Segmentation from Heterogeneous Datasets}

\textbf{Problem Statement:} Given a set of $K$ heterogeneous datasets $\{\mathcal{D}_k\}$, $1 \leq k \leq K$, let us formalize the joint learning segmentation problem. Each dataset $\mathcal{D}_k = \{(x, y)_n\}$ is composed of pairs $(x, y)_n$, where $x$ is an image and $y$ a segmentation mask assigning a label $l \in \mathcal{L}_k$ to every $i$-th voxel $x_i$. $\mathcal{L}_k$ is the labelset associated to dataset $\mathcal{D}_k$. We assume disjoint labelsets, except for the background label included in all datasets.
We aim at learning the parameters $\Theta$ for a single segmentation model $f(\hat{x}; \Theta)$ that, given a new image $\hat{x}$, produces a segmentation mask $\hat{y}$ where every voxel $\hat{y}_i \in \hat{\mathcal{L}} = \bigcup_{k = 1}^K \mathcal{L}_k$. The label space $\hat{\mathcal{L}}$ is built as the union of all labelsets, and we assign a single label to every voxel $\hat{y}_i$. 

\Enzo{Note that, since the new labelset $\hat{\mathcal{L}}$ includes all labels from all datasets, some structures that were labeled as background in one dataset may be labeled as foreground in other datasets, raising the label contradiction problem shown in Figures \ref{fig:intro} and \ref{fig:ace}.a. In these cases, the foreground labels (e.g. brain tissue labels) should prevail over the background labels in the final mask generated by the segmentation model.}

In case of MRI brain lesion and anatomy segmentation, we have $K=2$ brain MRI datasets. The first one, denoted $\mathcal{D_A}$, is annotated with anatomical (brain tissue) labels while the second one, referred as $\mathcal{D_L}$, considers brain lesions (tumor or WMH are the application scenarios studied in this work). The corresponding label spaces for every dataset are $\mathcal{L_A}$ and $\mathcal{L_L}$. In what follows, we describe multiple alternatives to train such model based on a standard U-Net architecture \cite{unet}.

\subsection{Naive Models}

We first consider a naive model where a single U-Net is trained by minimizing standard loss functions (typical categorical cross entropy and Dice losses), to perform joint learning from heterogeneous datasets. We employ a standard U-Net architecture (see Appendix \ref{app:networkDescription} for a complete description of the architecture) with a final \textit{softmax} layer producing $|\hat{\mathcal{L}}|$ probability maps, i.e. one for each class in the joint labelset $\hat{\mathcal{L}}$. Patch-based training is performed by constructing balanced mini-batches of image patches. We balance the mini-batches by sampling with equal probability from all datasets and all classes.

As stated in section \ref{sec:soa} and illustrated in Figure \ref{fig:ace}.a, labels coming from different datasets may contradict each other and mislead the training process. Brain tissue segmentations or cortical/sub-cortical structures generally cover the complete brain mass. However, lesion annotations like WMH and tumour cover only a small portion of it. The main issue with the proposed naive model arises from this fact: when sampling image patches containing small lesions, whatever is considered background in the patch should be actually classified as some type of brain tissue. However, since the lesion dataset does not contain brain tissue annotations, it will be considered as background. In other words, the model will be encouraged to classify brain tissue as background. In the results that will be presented in Section \ref{sec:results}, we provide empirical evidence of this issue and its impact in model performance.

\subsection{Multi-network Baseline}
A trivial solution to the aforementioned problem is to use multiple independent models, trained for every specific task. In this case, segmentation results are then combined following some kind of fusion scheme. In case of brain lesion and tissue segmentation, since lesion labels prevail over tissue labels, we can simply overwrite them. However, note that such model requires extra efforts at training time: we need to train a single model for every dataset, increasing not only the training time but also the overall model complexity, i.e. the number of learned parameters. Moreover, at test time, every model is evaluated on the test image and a label fusion strategy must be applied to combine the multiple predictions.

We consider a multi U-Net model as baseline to benchmark the proposed solution, training a single U-Net with categorical cross entropy in every dataset. Label fusion is implemented by overwriting the brain tissue segmentation with the (non-background) lesion masks.

\subsection{Adaptive Cross Entropy}
In this work, we propose to overcome the issues that arise when training a single CNN from heterogeneous (and potentially contradictory) datasets with a new loss function titled \textit{adaptive cross entropy} (ACE). Let us first recall the classical formulation of cross entropy. Given an estimate distribution $q$ for a true probability distribution $p$ defined over the same discrete set (in our setting, the set $\hat{\mathcal{L}}$ of possible labels, with $C = |\hat{\mathcal{L}}|$), the cross entropy between them is computed as:

\begin{equation}
	H(p, q) = - \sum_{j=1}^C p_j \cdot log(q_j).
\end{equation}

For a given voxel $x_i$ with ground-truth label $y_i \in \hat{\mathcal{L}}$ (with $ 1 \leq y_i \leq C = |\hat{\mathcal{L}}|$), we compute the categorical cross entropy loss between the voxel-wise model prediction $f(x_i; \Theta)$, and the corresponding one-hot encoded version of $y_i$ denoted by $e^{(y_i)}$ as:

\begin{align}
	H(x_i, y_i) &= - \sum_{j=1}^C e^{(y_i)}_j \cdot log(f(x_i; \Theta)_j) = - \sum_{j=1}^C \mathds{1}_{[y_i = j]} \cdot log(f(x_i; \Theta)_j) \nonumber\\
	&= - log(f(x_i; \Theta)_{y_i}).
\end{align}

The standard voxel-wise cross entropy loss $L_H$ is aggregated as the average loss considering all voxels $\{x_i\}_{1 \leq i \leq m}$ in the image patch: 

\begin{equation}
\label{eq:crossEntropy}
	L_H(x,y) = -\sum_{i=1}^m log(f(x_i; \Theta)_{y_i}).
\end{equation}

The cross entropy loss $L_H$ is minimized when the prediction equals the ground-truth. In the multi-task context discussed in this work, this raises the label contradiction problem between lesion background and brain tissue segmentation illustrated in Figure \ref{fig:ace}.a. This fact motivates the design of the \textit{adaptive cross entropy} (ACE) loss which behaves differently depending on the structures of interest under consideration. We reinterpret the meaning assigned to the background label of the lesion dataset as \textit{`any label that is not lesion'} and modify the loss function accordingly. The proposed \textit{adaptive cross entropy} is therefore defined as:
\begin{equation}
\label{eq:ACE}
	H^A(x_i, y_i)  = \begin{cases}
		- log(f(x_i; \Theta)_{y_i}) & \text{if } y_i \text{ \textbf{is not} lesion background} \\
		% - \sum_{c=1}^C \mathds{1}_{(\exists y_k) y_k = c} \cdot log(1 - f(x_i; \Theta)_{y_i}) & \text{si } y_i = -1 \\
		- log(\sum_{j \in \{\hat{\mathcal{L}} \setminus \mathcal{L}(y) \}} f(x_i; \Theta)_{j}) & \text{if } y_i \text{ \textbf{is} lesion background}
	\end{cases}
\end{equation}

\noindent where the set $\{\hat{\mathcal{L}} \setminus \mathcal{L}(y) \}$ contains all labels, except those in the current image patch ground-truth (referred as $\mathcal{L}(y)$). Equation \ref{eq:ACE} shows that ACE employs the standard cross entropy formulation when voxel $i$ is labeled as anything but lesion background. However, when voxel $i$ corresponds to lesion background, we compute  $-log(s)$, where $s = \sum_{j \in \{\hat{\mathcal{L}} \setminus \mathcal{L}(y) \}} f(x_i; \Theta)_{j}$ is the sum of scores $f(x_i; \Theta)_{j}$ for all classes $j$ that are not present in the patch $y$ (including background). In this way, when the label is not in conflict, minimizing $H^A$ is equivalent to maximizing the score for the correct class. However, when dealing with a voxel whose ground truth is lesion background (i.e. we are not sure about the brain tissue that corresponds to it), the model tends to maximize the probability for all non-lesion classes. Figure \ref{fig:ace}.b illustrates this idea. In practice, we compute the aggregated ACE loss $L_H^A$ for all voxels $\{x_i\}_{1 \leq i \leq m}$ in the image patch as: $L_H^A(x,y) = \frac{1}{m} \sum_{i=1}^{m} H^A(x_i, y_i).$

\Enzo{Note that in the ACE formulation, we sum over the scores before taking the logarithm. The reasoning behind having the sum inside the log function on the proposed adaptive cross entropy is to effectively unify those labels that are not lesion (i.e. background and brain tissue segmentations, which raise the label contradiction problem illustrated in Figure 2.a) in a unique class. We do that by assigning to this virtual class the sum of the scores the model assigned to each of those labels.}

Note that in the application scenarios studied in this work, lesion labels collide with brain tissues, motivating the ACE formulation given in Equation \ref{eq:ACE}. Nonetheless, given an arbitrary number of $K$ datasets, in general it is straightforward to apply the proposed ACE loss to different labels raising similar issues, by just changing the condition that adapts the loss behaviour.

\section{Experiments \& Results}
Six different datasets were used in the experimental comparative analysis. We consider joint learning of brain tissue segmentation and two separate type of lesions: brain tumor and WMH. We trained models specialized for brain tissue + WMH, and other models for brain tissue + tumor, showing that the proposed ACE loss function can generalize to different scenarios.
\begin{figure}[t!]
    \label{fig:results}
	\includegraphics[width=1\textwidth]{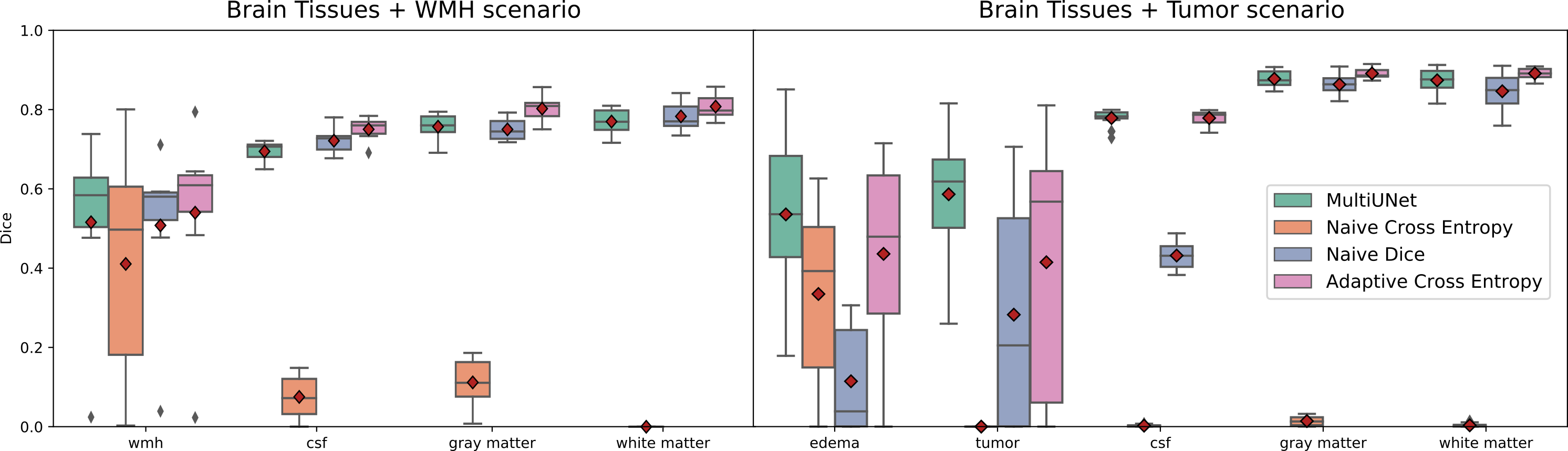}
	\caption{Experimental results obtained when comparing a single model trained with the proposed ACE loss, with the Multi-UNet and the naive cross entropy and Dice models (red diamond indicates the mean value). Note that a single model trained with ACE achieves equivalent performance to that of Multi-UNet, while naive models under-perform by a big margin in both cases.}
\end{figure}

% \usepackage{multirow}
% \usepackage{graphicx}
\begin{table}[]
\centering
\resizebox{\textwidth}{!}{%
\begin{tabular}{|l|c|c|c|c|c|c|c|c|c|c|c|c|c|c|c|c|c|c|}
\hline
\multirow{3}{*}{} & \multicolumn{8}{c|}{\textbf{Brain Tissues + WMH}} & \multicolumn{10}{c|}{\textbf{Brain Tissues + Tumor}} \\ \cline{2-19} 
 & \multicolumn{2}{c|}{\textbf{WMH}} & \multicolumn{2}{c|}{\textbf{CSF}} & \multicolumn{2}{c|}{\textbf{GM}} & \multicolumn{2}{c|}{\textbf{WM}} & \multicolumn{2}{c|}{\textbf{Edema}} & \multicolumn{2}{c|}{\textbf{Tumor}} & \multicolumn{2}{c|}{\textbf{CSF}} & \multicolumn{2}{c|}{\textbf{GM}} & \multicolumn{2}{c|}{\textbf{WM}} \\ \cline{2-19} 
 & Mean & Std & Mean & Std & Mean & Std & Mean & Std & Mean & Std & Mean & Std & Mean & Std & Mean & Std & Mean & Std \\ \hline
\textbf{Multi UNet} & 0.516 & 0.232 & 0.694 & 0.028 & 0.757 & 0.035 & 0.77 & 0.035 & 0.509 & 0.228 & 0.586 & 0.143 & 0.778 & 0.021 & 0.877 & 0.02 & 0.874 & 0.026 \\ \hline
\textbf{Naive CE} & 0.411 & 0.294 & 0.075 & 0.057 & 0.112 & 0.067 & 0 & 0 & 0.335 & 0.219 & 0 & 0 & 0.002 & 0.003 & 0.013 & 0.011 & 0.003 & 0.004 \\ \hline
\textbf{Naive Dice} & 0.508 & 0.218 & 0.721 & 0.035 & 0.75 & 0.029 & 0.783 & 0.038 & 0.114 & 0.126 & 0.282 & 0.252 & 0.432 & 0.032 & 0.863 & 0.025 & 0.846 & 0.042 \\ \hline
\textbf{ACE} & 0.54 & 0.245 & 0.75 & 0.031 & 0.802 & 0.034 & 0.807 & 0.033 & 0.414 & 0.264 & 0.415 & 0.3 & 0.779 & 0.018 & 0.891 & 0.013 & 0.891 & 0.012 \\ \hline
\end{tabular}%
}
\caption{Numerical results corresponding to the experiments shown in Figure \ref{fig:results}.}
\label{tab:results}
\end{table}

\subsubsection*{Brain tissues + WMH scenario}
We employed the training data provided by the MRBrainS13 Challenge \cite{mendrik2015mrbrains}  (brain tissue annotations), the WMH Segmentation Challenge \cite{mrbrains17} (WMH lesions) and MRBrains18 \cite{mrbrains18} (brain tissues + WMH). We trained/validated our models using the training partition of MRBrainS13 as anatomical dataset ($\mathcal{D_A}$) and WMH Segmentation Challenge as lesion dataset ($\mathcal{D_L}$). For testing, we used the joint segmentations provided for training in the MRBrainS2018 Challenge, to evaluate the simultaneous predictions. The data from the MRBrainS13 Challenge consists of 5 images with brain tissue annotations, of which 4 were used for training, and the remaining one for validation. The WMH Segmentation Challenge provides 60 images with the corresponding WMH reference segmentation, of which 48 were used for training, and the rest for validation. The MRBrainS18 Challenge provides 7 images, which were all used for evaluation.

\subsubsection*{Brain tissues + Tumor scenario}
Given the lack of datasets with simultaneous annotations for brain tumors and tissues, we resorted to using synthetic and simulated images. We trained/validated our models using 15 images from the Brainweb \cite{cocosco1997brainweb} synthetic brain phantoms with brain tissue annotations for the anatomical dataset ($\mathcal{D_A}$). For the lesion dataset ($\mathcal{D_L}$) we employed 50 simulated tumor images available from the BRATS2012 challenge \cite{brats12}. For testing, we simulated 20 brain tumors using Tumorsim \cite{prastawa2009tumorsim}, using 5 healthy Brainweb phantom probability maps. In that way, combined segmentations of brain tissue and tumors were available for testing. Note that, for the sake of fairness, healthy images used to simulate brain tumors for testing were not included in the training dataset ($\mathcal{D_A}$). 
\label{sec:results}

\begin{figure}[t!]
\includegraphics[width=1\linewidth]{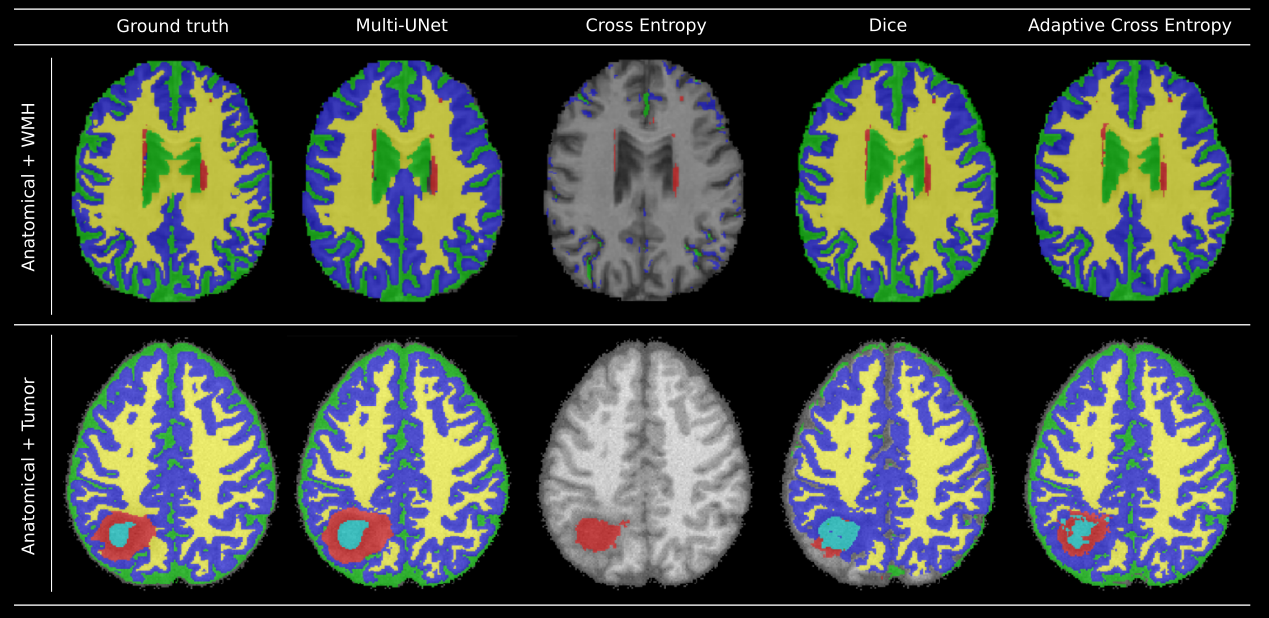}
\caption{Qualitative results for both scenarios (brain tissues + WMH in the top row, and brain tissues + tumor segmentation in the bottom row). Note that using naive cross entropy and Dice losses result in very poor performance. The proposed ACE makes it possible to train a single model for both tasks with equivalent performance to multiple networks by solving the label contradiction issues.}
\label{fig:qualitativeComparison}
\end{figure}

\subsubsection*{Results \& Discussion}
Figure \ref{fig:results} summarizes the quantitative results for both application scenarios, when comparing the Multi-UNet model with single models trained with naive cross entropy and Dice functions as well as the proposed ACE\footnote{We implemented the CNN in Keras and trained it using Adam optimizer with default parameters. Balanced mini-batches of 7 image patches of size $32 \times 32 \times 32$ are used during training. A complete description of the baseline UNet architecture used for both, single and multi-network models, is provided in Appendix \ref{app:networkDescription}.} (see Figure \ref{fig:qualitativeComparison} for qualitative results). As expected, the Multi-UNet model trained with standard cross entropy outperforms the single models trained with naive losses. More importantly, our proposed ACE makes it possible to train a single model for joint learning of brain lesion and anatomy from heterogeneous datasets, achieving equivalent performance to that of Multi-UNet. 

This is due to the fact that both, Multi-UNet and the single ACE models, are not affected by the label contradiction problem illustrated in Figure \ref{fig:ace}.a.  Note that in case of brain tissue segmentation, the single model trained with ACE tends to outperform even the Multi-UNet model. As discussed in \cite{rajchl2018neuronet}, learning jointly from hierarchical sets of class labels has the potential to increase the overall accuracy based on theory derived from multi-task learning. We hypothesize that this increase in performance is related to this fact: since the model trained with ACE learns to predict lesion and tissues simultaneously, it can also learn label interactions that the Multi-UNet can not capture. 

\Enzo{A deeper analysis of the quantitative results reveals that the single UNet model trained with the proposed ACE achieved equivalent performance to the Multi-UNet in WMH segmentation (no significant differences according to Wilcoxon test), better or equivalent performance in terms of brain tissue segmentation (depending on the brain structure) and only worse performance for edema and tumor. This worse performance for edema and tumor is explained by the fact that the Multi-UNet was trained using all available modalities per dataset, while the single UNet was trained using only those modalities available in both, anatomical and lesion datasets. This is a limitation of our approach when compared with multiple UNets trained for specific tasks: since we perform joint training of a single model with fixed number of input channels, we can only use those sequences available in both anatomy and lesion datasets. In case of edema and brain tumor segmentation, the Multi-UNet was trained with multiple MR modalities for the tumor segmentation task (it uses T1, T1g, T2 and FLAIR) while the single UNet was trained using only T1 images (all details about available MR modalities for every dataset are provided in Appendix \ref{app:seqs}). This requirement may represent a limitation if the datasets depend on different types of image modalities. There are alternatives that could be considered to deal with this issue like imputing the missing modalities by means of image synthesis or using ad-hoc techniques like the HeMIS (Hetero-Modal Image Segmentation) model by \cite{havaei2016hemis}.}

Even if all images used in the experiments are MRI, there is a shift in the distribution of image intensities when we go from datasets used at training and test time. This is known as the multi-domain problem, and is usually addressed using domain adaptation techniques \cite{kamnitsas2017unsupervised}. In this work, we did not take into account the multi-domain problem. In the future, we plan to extend the proposed method and incorporate domain adaptation, further improving the accuracy of the results.

\section{Conclusions}
In this work we proposed the \textit{adaptive cross entropy} loss, a novel function to perform joint learning of brain lesion and anatomy segmentation from heterogeneous datasets using CNNs. The proposed loss takes into account potential label contradiction conflicts that can arise when training segmentation algorithms for multiple tasks using datasets with disjoint annotations. We trained single CNN models using the proposed ACE, naive cross entropy and Dice losses, and compared their performance with a Multi-UNet model where independent CNNs were trained for every task. Experimental evaluations in two scenarios provided empirical evidence about the effectiveness of the proposed approach.

In the future, we plan to extend the evaluation of the proposed loss function \Enzo{to other CNN architectures (Deepmedic \cite{deepmedic} for example) and} to alternative brain MRI segmentation scenarios (e.g. considering subcortical structures as anatomical segmentation or traumatic brain injuries as lesions). Moreover, we plan to investigate the effects of the multi-domain problem in this context, and incorporate domain adaptation strategies to address this issue when learning from heterogeneous datasets.

\Enzo{Regarding the ACE formulation, we plan to explore alternative weighting mechanisms within the loss function that could help to alleviate the class-imbalance problems that could emerge when dealing with tiny structures of interest.}

%\begin{table}[htbp]
 % The first argument is the label.
 % The caption goes in the second argument, and the table contents
 % go in the third argument.
%\floatconts
%  {tab:example}%
%  {\caption{An Example Table}}%
%  {\begin{tabular}{ll}
%  \bfseries Dataset & \bfseries Result\\
%  Data1 & 0.12345\\
%  Data2 & 0.67890\\
%  Data3 & 0.54321\\
%  Data4 & 0.09876
%  \end{tabular}}
%\end{table}

\newpage
% Acknowledgments---Will not appear in anonymized version
\midlacknowledgments{NR is now at Google. EF is beneficiary of an AXA Research Grant. We thank NVIDIA Corporation for the donation of the Titan X GPU used for this project. DFS is partially supported by Universidad de Buenos Aires and CONICET.}

\bibliography{roulet19}

\appendix

\section{Detailed Network Architecture}
The architecture used in this work is based on a standard U-Net \cite{unet}. It can be divided into a contraction and an expansion path. Each path is a sequence of four convolution blocks, composed of two convolutional layers with $3 \times 3 \times 3$ kernels and one voxel of padding, each one followed a ReLU activation layer. We also used batch-normalization to ease training. Every block from the contraction path is connected to the next one by a $2 \times 2 \times 2$ max-pooling layer, while the blocks from the expansion path are connected by $2 \times 2 \times 2$ transposed convolutions for upsampling. The output from each block of the contraction path is added to the input of the corresponding block from the expansion path to combine the localized features of the former with the high level information from the latter. This is in contrast with standard U-Net which uses concatenation of feature maps instead of sumation. The layers from the first block have 32 channels. The number of channels is doubled in every max-pooling layer and halved in every transposed convolution layer.
Finally, a $1 \times 1 \times 1$ convolution layer with softmax activation is used to convert the output of the last layer into voxel-wise label probability maps. 

We implemented the CNN in Tensorflow and trained it using Adam optimizer. The weights were initialized using He method \cite{he2015delving}. Balanced mini-batches of 7 image patches of size $32 \times 32 \times 32$ were used during training. 
\label{app:networkDescription}

\section{MR Sequences Available Per Dataset}
\label{app:seqs}
Different MR sequences were available for every dataset. Table \ref{tab:sequences} summarizes this information. 

% \usepackage{multirow}
% \usepackage{graphicx}
\begin{table}[H]
\resizebox{\textwidth}{!}{%
\begin{tabular}{|c|c|c|c|c|c|c|}
\hline
Scenario                              & Dataset    & T1 & T1 with Gadolinium (T1g) & T2 & IR & FLAIR \\ \hline
\multirow{3}{*}{Brain Tissue + WMH}   & MRBrains13 & X  &                    &    & X  & X     \\ \cline{2-7} 
                                      & WMH        & X  &                    &    &    & X     \\ \cline{2-7} 
                                      & MRBrains18 & X  &                    &    & X  & X     \\ \hline
\multirow{3}{*}{Brain Tissue + Tumor} & BrainWeb   & X  &                    &    &    &       \\ \cline{2-7} 
                                      & BRATS12    & X  & X                  & X  &    & X     \\ \cline{2-7} 
                                      & Tumorsim   & X  & X                  & X  &    & X     \\ \hline
\end{tabular}%
}
\caption{MR sequences available per dataset.}
\label{tab:sequences}
\end{table}

The UNet architecture used in our experiments can receive multiple MR sequences as input by simply interpreting them as multiple image channels. Note that the Multi-UNet network was trained with as many sequences as possible per task. For example, if T1, T2 and FLAIR sequences were available in the lesion dataset and only T1, T2 were available for the anatomy dataset, we trained every independent UNet using all available sequences (of course, these sequences have to be available in the test dataset as well). However, when training the single UNet models using the naive losses and ACE, we can only use those sequences available in both anatomy and lesion datasets.

Given the MR sequences available for every dataset (shown in Table \ref{tab:sequences}) we trained the single and multi-network models under the following setting:
\begin{itemize}
    \item Brain Tissue + WMH scenario: The Multi-UNet model was trained and tested using T1+IR+FLAIR for the brain tissue segmentation task, and T1+FLAIR for the WMH segmentation task. The single UNet models were trained using only T1+FLAIR for all tasks.
    \item Brain Tissue + Tumor scenario: The Multi-UNet model was trained and tested using T1 for the brain tissue segmentation task, and  T1+T1g+T2+FLAIR for the tumor segmentation task. The single UNet models were trained using only T1 for all tasks.
\end{itemize}

Note that this setting gives some advantages to the Multi-UNet model over the single model trained with ACE, since it uses more MR sequences for the lesion segmentation task. This is reflected in the results shown in Figure \ref{fig:results}, specially for the brain lesion segmentation task, where the better performance shown by the Multi-UNet model with respecto to the single model trained with ACE can be explained by this difference in the number of sequences used to train them.
\end{document}